\pdfoutput=1

\documentclass[11pt]{article}

\usepackage[final]{acl}

\usepackage{times}
\usepackage{latexsym}

\usepackage[T1]{fontenc}

\usepackage[utf8]{inputenc}

\usepackage{microtype}

\usepackage{inconsolata}

\usepackage{graphicx}

\usepackage[utf8]{inputenc} %
\usepackage[T1]{fontenc}    %
\usepackage{hyperref}       %
\usepackage{url}            %
\usepackage{booktabs}       %
\usepackage{amsfonts}       %
\usepackage{nicefrac}       %
\usepackage{microtype}      %
\usepackage{xcolor}         %

\usepackage{enumitem} 
\usepackage{pifont}   %

\usepackage{amsthm}

 \usepackage [ruled]{algorithm2e}
\usepackage{lipsum}
\usepackage{graphicx}
\usepackage{wrapfig}
\usepackage[size=small]{caption}
\usepackage{mathtools}
\usepackage{amssymb}
\usepackage{amsthm}
\usepackage{soul}
\usepackage[export]{adjustbox}
\usepackage{pifont}
\usepackage{makecell}
\usepackage{tabularx}
\usepackage{multirow}
\usepackage[capitalize,noabbrev]{cleveref}
\definecolor{mark}{RGB}{208,64,56}

\definecolor{darkred}{rgb}{0.0, 0.0, 0.55}

\usepackage{lineno}
\usepackage{xcolor,colortbl}

\usepackage[export]{adjustbox}
\definecolor{green}{HTML}{009B55}
\definecolor{customgray}{rgb}{0.25,0.25,0.25}
\definecolor{customred}{rgb}{0.8,0.05,0.05}
\definecolor{urlcolors}{rgb}{0.872,0.2,0.552}

\newcommand{\rnw}{\textcolor{green}{\ding{51}} + \textcolor{red}{\ding{55}}}
\newcommand{\onlyr}{\textcolor{green}{\ding{51}}}
\newcommand{\onlyw}{\textcolor{red}{\ding{55}}}

\newcommand{\method}{\texttt{W2SR}\xspace}
\newcommand{\methodp}{\texttt{W2SR-P}\xspace}
\newcommand{\methodn}{\texttt{W2SR-N}\xspace}

\newcommand{\gray}{\cellcolor{gray!20}}
\newcommand{\blue}{\cellcolor{blue!6}}

\usepackage[most,skins,theorems]{tcolorbox}
\usepackage{tikz}
\tcbset{
  aibox/.style={
    width=\linewidth,
    top=8pt,
    bottom=4pt,
    colback=blue!6!white,
    colframe=black,
    colbacktitle=black,
    enhanced,
    center,
    attach boxed title to top left={yshift=-0.1in,xshift=0.15in},
    boxed title style={boxrule=0pt,colframe=white,},
  }
}

\newtcolorbox{AIbox}[2][]{aibox,title=#2,#1}

\usepackage{contour}

\title{Incentivizing Strong Reasoning from Weak Supervision}

\author{%
Yige Yuan$^{1,2,5}$\thanks{Work done during an internship at Alibaba Group.}, \quad Teng Xiao$^{3,4}$, \quad Shuchang Tao$^{5}$, \\ \textbf{Xue Wang}$^{5}$\textbf{,}  \quad \textbf{Jinyang Gao}$^{5}$\textbf{,}  \quad\textbf{Bolin Ding}$^{5}$\textbf{,} \quad \textbf{Bingbing Xu}$^{1}$\thanks{Corresponding author.} \\
$^1$Institute of Computing Technology, Chinese Academy of Sciences\\
$^2$University of Chinese Academy of Sciences, \\
$^3$University of Washington, 
$^4$Allen Institute for AI,
$^5$Alibaba Group \\
\texttt{yuanyige20z@ict.ac.cn, tengxiao01@gmail.com, shuchangtao5@gmail.com}
}

\begin{document}
\maketitle

\begin{abstract}
Large language models (LLMs) have demonstrated impressive performance on reasoning-intensive tasks, but enhancing their reasoning abilities typically relies on either reinforcement learning (RL) with verifiable signals or supervised fine-tuning (SFT) with high-quality long chain-of-thought (CoT) demonstrations, both of which are expensive. In this paper, we study a novel problem of incentivizing the reasoning capacity of LLMs without expensive high-quality demonstrations and reinforcement learning. 
We investigate whether the reasoning capabilities of LLMs can be effectively incentivized via supervision from significantly weaker models. We further analyze when and why such weak supervision succeeds in eliciting reasoning abilities in stronger models. Our findings show that supervision from significantly weaker reasoners can substantially improve student reasoning performance, recovering close to 94\% of the gains of expensive RL at a fraction of the cost. Experiments across diverse benchmarks and model architectures demonstrate that weak reasoners can effectively incentivize reasoning in stronger student models, consistently improving performance across a wide range of reasoning tasks. Our results suggest that this simple weak-to-strong paradigm is a promising and generalizable alternative to costly methods for incentivizing strong reasoning capabilities at inference-time in LLMs. Code is publicly available at \href{https://github.com/yuanyige/w2sr}{\color{customred}{this link}}.

\begin{figure*}[!h]
    \centering
    \includegraphics[width=0.98\linewidth]{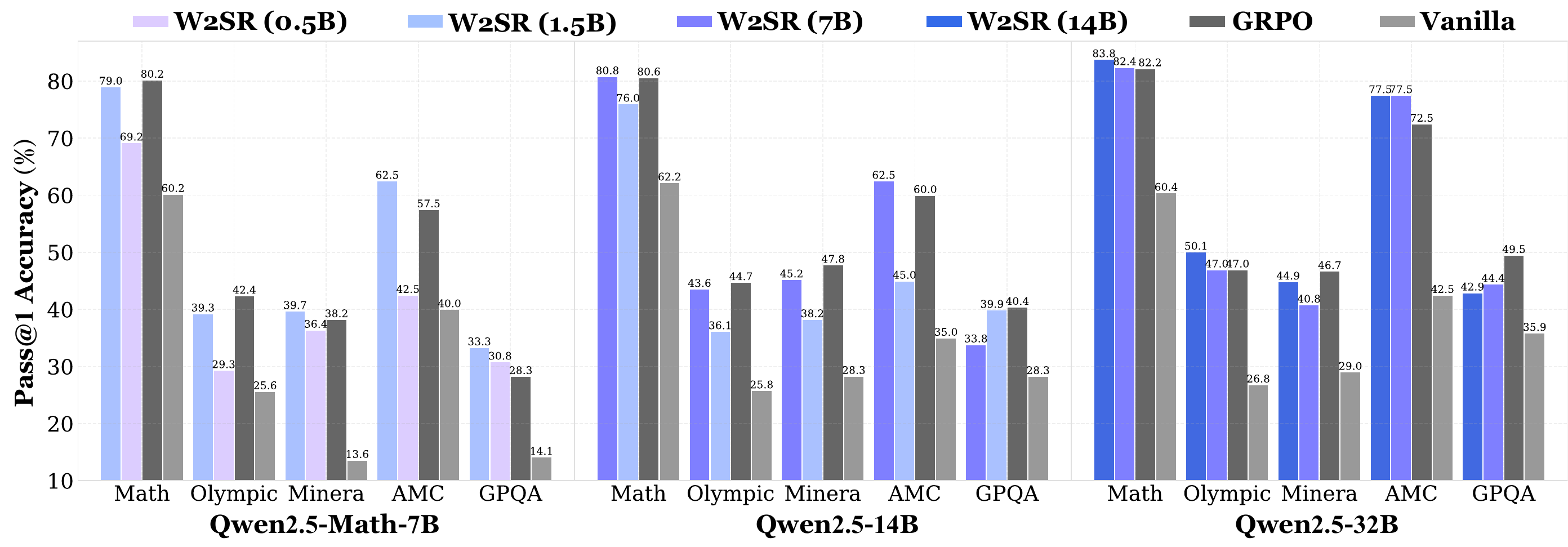}
    \caption{
    Benchmark performance of \method  across student scales (Qwen2.5-Math-7B, Qwen2.5-14B, and Qwen2.5-32B). 
Each student model is trained with weak teachers: 7B uses 0.5B/1.5B teachers, 14B uses 1.5B/7B teachers, and 32B uses 7B/14B teachers. 
We compare \method to GRPO and vanilla models on five reasoning benchmarks. 
Weak supervision consistently yields strong reasoning, rivaling RL with far lower cost.
    } 
    \label{fig:motivation}
    \vspace{-5mm}
\end{figure*}
\end{abstract}

\section{Introduction}
\label{sec:intro}

Large language models (LLMs) have demonstrated strong performance across a variety of reasoning-intensive tasks, such as mathematical problem solving~\cite{guo2025deepseek,yang2024qwen2}, symbolic computation~\cite{fang2024large}, and code generation~\cite{jiang2024survey}, often achieving results competitive with or even surpassing human-level capabilities~\cite{xiao2025simper,NEURIPS2024_cf8b2205,xiao2025on,xiao-etal-2024-leverage}. A key technique for enabling such reasoning abilities is to scale up inference compute via long chain-of-thought (CoT), that encourages models to explicitly generate intermediate reasoning steps before arriving at a final answer~\cite{guo2025deepseek,team2025kimi,wei2022chain}. This approach has proven effective in improving answer accuracy and enabling the decomposition of complex problems into more manageable subproblems.

A mainstream way to incentivize the reasoning ability of LLMs is \textit{reinforcement learning } (RL) with verifiable reward signals, which shows strongest empirical gains but at the price of substantial compute and data engineering~\cite{guo2025deepseek,team2025kimi}.  
In addition to the high computational cost of thousands of GPU‑hours per run, RL is effective only when the base model can discover correct trajectories during roll‑outs, yet many open-source models fail to meet this assumption. Recent studies further indicate that while RL with verifiable reward can boost sampling efficiency, it may simultaneously limit the
exploration capacity, resulting in a
narrower reasoning capability boundary compared to base models~\cite{shao2024deepseekmath,yue2025does}. 
A more compute-friendly alternative strategy is \textit{supervised fine‑tuning} (SFT) on high‑quality chain‑of‑thought (CoT) data distilled from a teacher model~\cite{yeo2025demystifying,ye2025limo,muennighoff2025s1}. 
However, collecting high-quality CoT data remains challenging in specialized domains, where human evaluation is time-consuming or costly. Obtaining reasoning demonstrations from frontier (i.e., extremely large) teacher models can also be prohibitively expensive, especially at scale. Moreover, the assumption of high-quality supervision from stronger models, may not hold for superhuman models~\cite{burnsweak}. This persistent trade‑off remains unsolved and calls for a fresh perspective for incentivizing reasoning.

As an alternative, in this work, we ask a fundamental yet unexplored question:
\emph{can we incentivize the reasoning capacity of LLMs without expensive strong teacher models and reinforcement learning?}

To address this question, we investigate whether the reasoning capacities of LLMs can be effectively incentivized through supervision from significantly weaker models. Specifically, we propose a lightweight and scalable Weak-to-Strong Reasoning paradigm (\method), where a strong student model is trained on CoT trajectories generated by much weaker teacher models. \kern-0.2emOur hypothesis is that, even if a weak reasoner underperforms relative to the student or is considerably smaller in size, it can still provide imperfect yet informative reasoning traces that help elicit the student's reasoning abilities.

To investigate our hypothesis, we conduct extensive empirical studies using supervised fine-tuning (SFT) on strong student models with long CoTs distilled from weak teachers. As shown in~\cref{fig:motivation}, our analysis across multiple reasoning benchmarks reveals that weak supervision can incentivize strong reasoning abilities. For example, distilling reasoning traces from a \texttt{0.5B–1.5B} teacher enables a \texttt{7B–32B} student to recover up to 94.34\% of the reasoning gains from reinforcement learning (e.g., +18.8 Pass@1 on \textsc{MATH}), outperforming both the teacher and, in some cases, high-cost RL baselines. We further investigate when weak supervision is most effective. Results show that teacher reasoning ability, such as producing structurally well-formed CoTs, is more important than model size or final accuracy. Surprisingly, even imperfect or partially incorrect traces can incentivize reasoning in the student, while increasing teacher strength beyond a moderate level yields diminishing returns.

Our findings advocate a lightweight and scalable paradigm for
incentivizing reasoning LLMs.
Instead of collecting high-quality CoTs form expensive strong teachers or running costly RL,
we distill \emph{coarse yet structured} CoTs from much weaker supervision, which is sufficient to unlock the student’s latent reasoning circuits.
To sum up, our contributions are three‑fold:
(i)~We \emph{formulate and validate} weak‑to‑strong reasoning distillation,
demonstrating that strong reasoning abilities can
be incentivized from weak supervision to $4\times$ smaller and less accurate.
(ii)~Through Extensive ablations and analysis, we reveal \emph{the surprisingly key to effective supervision}: the teacher's reasoning ability rather than the model size or performance.
(iii)~We demonstrate \emph{practical benefits}:
The new paradigm achieves drastic cost reduction compared to both SFT and RL, while outperforming the teacher model and even high-cost RL baselines in some cases, providing a privacy-friendly path for domain experts to refine frontier models using lightweight local teachers.
Together, these contributions position weak‑to‑strong distillation as a promising paradigm for the lightweight and widely accessible reasoning LLMs.

\section{Related Work}
\label{sec:background}
\subsection{Large Reasoning Models}
Large language models (LLMs) have demonstrated profound capabilities in many domains. A recent line of research aims to enhance reasoning capabilities through inference-time scaling, as demonstrated by OpenAI’s o1\cite{jaech2024openai}, DeepSeek’s R1\cite{guo2025deepseek}, and Kimi K1.5~\cite{team2025kimi}. These methods typically prompt models to generate ultra-long reasoning traces to solve complex problems. At the training stage, reasoning abilities are generally improved using two main strategies: (1) reinforcement learning with verifiable rewards (RLVR)~\cite{guo2025deepseek,lambert2024t}, and (2) supervised fine-tuning (SFT) on human-curated or model-distilled data\cite{guo2025deepseek,ye2025limo,muennighoff2025s1}. While both approaches have shown promise, each comes with notable limitations. In particular, RL-based fine-tuning is significantly more computationally intensive, often requiring thousands of GPU-hours per run. A more compute-efficient alternative is to apply SFT on high-quality chain-of-thought (CoT) data distilled from a stronger teacher model~\cite{ye2025limo,muennighoff2025s1,guo2025deepseek}. However, collecting such data remains challenging, especially in specialized domains where human annotation is costly and time-consuming. Moreover, extracting reasoning traces from frontier (extremely strong) teacher models incurs substantial computational overhead, especially when scaling to large datasets, and may be infeasible if the model is already highly capable. \emph{In contrast to prior work, we shift focus to a novel and underexplored direction: understanding and leveraging the reasoning patterns of weaker models within a weak-to-strong generalization paradigm. Our work investigates whether and how weak supervision can incentivize strong reasoning capabilities.}

\subsection{Weak to Strong Generalization}
Weak-to-strong generalization refers to scenarios where stronger models surpass their weaker supervisors after fine-tuning~\cite{burnsweak}. This paradigm has emerged as a promising framework for addressing the challenges of AI alignment, particularly in the context of superalignment~\cite{leike2018scalable}, where future AI systems may exceed human capabilities, rendering human supervision inherently weak or insufficient. 
The weak-to-strong approach leverages weaker models to guide the training of stronger models, with the potential to unlock advanced capabilities while maintaining alignment with human values. 
This framework has been extensively explored through algorithmic innovations~\cite{liu2024co,guo2024improving}, empirical studies~\cite{tao2024your,yang2024weak,ye2024weak}, and theoretical analyses~\cite{lang2024theoretical}. \citep{yang2024weak} studies reasoning tasks within the weak-to-strong learning framework. However, their approach still relies on supervised fine-tuning using a selectively curated high-quality dataset. Moreover, prior work has largely focused on relatively simple tasks without requiring long CoT, leaving open the question of whether and how strong reasoning with long CoT can be effectively induced through weak supervision. In this paper, we present the first  study that fine-tunes LLMs using long CoT trajectories generated by significantly weaker reasoners, and demonstrate that such weak supervision can substantially enhance general reasoning abilities. \emph{Our approach is orthogonal to existing strategies, providing a compute-efficient and scalable alternative for incentivizing strong reasoning capabilities with inference-time scaling.}

Due to space constraints, we provide additional related works and discussions in Appendix~\ref{app:related}.

\section{Incentivizing Strong Reasoning Capacity from Weak Supervision}
\label{sec:method}

\subsection{Weak-to-Strong Reasoning Paradigm}

Let $\mathbf{x} = \{\mathbf{x}_i\}_{i=1}^{N}$ denote a set of $N$ input questions, where each question $\mathbf{x}_i = (x^1, x^2, \dots, x^T)$ is represented as a sequence of $T$ input tokens. The corresponding ground-truth answers are given by $\mathbf{a} = \{\mathbf{a}_i\}_{i=1}^{N}$.  
LLM policy $\pi_\theta$ generates output sequences $\mathbf{y} = \{\mathbf{y}_i\}_{i=1}^{N}$. For each output sequence $\mathbf{y}_i$, the model defines a conditional probability distribution over tokens $\pi_\theta(y^t \mid \mathbf{x}, \mathbf{y}^{<t})$, where $y^t$ is the token at position $t$, and $\mathbf{y}^{<t}$ denotes the preceding tokens. From each $\mathbf{y}_i$, we extract a substring $\hat{\mathbf{a}}_i \subseteq \mathbf{y}_i$ as the predicted answer for question $i$. We define the CoT as a subsequence of $\mathbf{y}_i$ that contains intermediate reasoning steps (i.e., reasoning trajectories) leading to the final answer $\hat{\mathbf{a}}_i$.

In this work, we propose a simple weak-to-strong reasoning (\method) paradigm. For each input question $\mathbf{x}_i$, a weak reasoner (the \emph{teacher}) generates a long CoT trajectory $\mathbf{y}_i$, which we use as weak supervision to fine-tune a stronger student model $\pi_{\theta_s}$ that initially lacks explicit incentivization for reasoning. The student is trained via simple SFT to imitate the teacher's reasoning trajectories, despite their potential imperfections. The goal of the \method paradigm is to incentivize stronger reasoning capabilities in the student through weak yet structured supervision provided by the teacher.

Note that we define a \emph{weak reasoner} as a teacher model with \emph{explicitly incentivized but limited reasoning abilities}, typically due to smaller model size, lower answer accuracy, or both.  Conversely, a \emph{strong student} is a model with greater capacity or better performance relative to its teacher. 
We refer to teacher models with explicitly incentivized reasoning capabilities as \texttt{-Reasoner} or \texttt{(R)}, and those without such capabilities as \texttt{-Non-Reasoner}.
To  evaluate the effectiveness of our \method{} paradigm, we investigate three \emph{weak-to-strong} variants based on the correctness of the teacher's final answer:
\vspace{-1mm}
\begin{itemize}[leftmargin=1.5em]
    \item \method: Uses all CoT trajectories from the weak teacher, regardless of answer correctness:
    \begin{equation}
        \mathcal{D} = \{(\mathbf{x}_i, \mathbf{y}_i)\}_{i=1}^N, \; \text{where} \; \mathbf{y}_i = \pi_{\theta_t}(\mathbf{x}_i).
    \end{equation}
    
    \item \methodp: Uses only CoT trajectories that yield the correct final answer from the weak teacher:
    \begin{equation}
        \mathcal{D}_p = \{(\mathbf{x}_i, \mathbf{y}_i) \mid \hat{\mathbf{a}}_i = \mathbf{a}_i \}.
    \end{equation}
    
    \item \methodn: Uses only trajectories  with incorrect final answers from the weak teacher:
    \begin{equation}
        \mathcal{D}_n = \{(\mathbf{x}_i, \mathbf{y}_i) \mid \hat{\mathbf{a}}_i \neq \mathbf{a}_i \}.
    \end{equation}
\end{itemize}
\vspace{-2mm}

The \emph{student} model $\pi_{\theta_s}$ is fine-tuned on the above CoT data via simple SFT, which updates its parameters $\theta_s$ by minimizing the following negative log-likelihood (NLL) loss.
\vspace{-2mm}
\begin{equation}
\mathcal{L}_{\text{SFT}}(\theta_s) = -\sum_{(\mathbf{x}, \mathbf{y}) \in \mathcal{D}} \sum_{t=1}^{|\mathbf{y}|} \log \pi_{\theta_s}(y^t \mid \mathbf{x},  \mathbf{y}^{<t})
\vspace{-2mm}
\end{equation}

\subsection{Evaluation via Reasoning Gap Recovered}

To evaluate the effectiveness of the \method paradigm, we introduce a metric called Reasoning Gap Recovered (RGR). RGR measures how much of the reasoning improvement achieved through RL can be recovered by supervising a strong student with a weak reasoner. It quantifies the extent to which weak-to-strong training closes the performance gap relative to a student whose reasoning capabilities are incentivized via computationally intensive RL. Specifically, we define RGR based on:
\begin{itemize}[leftmargin=1em]
\item \texttt{Weak Reasoner}: The weak teacher model with reasoning capability incentivized via RL.

\item \texttt{Weak-to-Strong Reasoner} (\method): Stronger student model trained via SFT on noisy or imperfect CoT traces produced by weak reasoner.

\item \texttt{Strong Reasoner}: The same strong student trained with RL, serving as performance ceiling.
\end{itemize}

The following RGR metric quantifies the performance gap of the Pass@1 score between a weak reasoner and two strong reasoners that are trained with weak supervision and reinforcement learning (RL), respectively. Higher RGR indicates that weak supervision from the weak reasoner effectively recovers or even exceeds the performance of the RL-trained strong reasoner. In contrast, lower RGR suggests that weak-to-strong training provides only limited improvement over the weak reasoner.
\begin{equation}
\texttt{\small RGR} = \frac{\texttt{\small Weak-to-Strong Reasoner} \;-\; \texttt{\small Weak Reasoner}}{\texttt{\small Strong Reasoner} \;-\; \texttt{\small Weak Reasoner}}.
\end{equation}

Note that RGR can exceed 1 when the student trained under weak supervision outperforms its RL-trained counterpart, demonstrating that structured but imperfect traces from weaker models can sometimes elicit superior reasoning. Compared to raw accuracy, RGR normalizes performance gains relative to the RL upper bound, offering a faithful measure of reasoning-specific transfer.

\begin{table*}[!t]

\centering
\caption{Weak-to-Strong Reasoning Performance Comparison. Evaluation of three student models (7B–32B) trained with supervision from four weaker reasoner teachers (0.5B–14B) across five reasoning benchmarks. Reports Metrics of Pass@1 (\%) and Reasoning Gap Recovered (RGR \%) for (1) base teacher, (2) base student, as well as student trained with (3) GRPO, (4) LIMO and (5) our Weak-to-Strong Reasoning (\method, \methodp). ``-'' indicates RGR is inapplicable. Best performance is marked with \textbf{boldface}, second best of our method is marked with \underline{underline}. (R) denotes the teacher possesses reasoning capabilities.}
\vspace{-2mm}
\begin{adjustbox}{width=0.98\textwidth}
\setlength{\tabcolsep}{0.2mm}
\small
\label{tab:weak}
\begin{tabular}{lll|cccccccccccc}
\toprule[1.5pt]
\textbf{Datasets ($\rightarrow$)} & \textbf{Metrics ($\rightarrow$)}& & \multicolumn{2}{c}{\textbf{Math}} & \multicolumn{2}{c}{\textbf{Olympic}} & \multicolumn{2}{c}{\textbf{Minera}} & \multicolumn{2}{c}{\textbf{AMC}} & \multicolumn{2}{c}{\textbf{GPQA}} & \multicolumn{2}{c}{\textbf{Average}} \\
\cmidrule(lr){4-5} \cmidrule(lr){6-7} \cmidrule(lr){8-9} \cmidrule(lr){10-11} \cmidrule(lr){12-13} \cmidrule(lr){14-15} 
\textbf{Student ($\downarrow$)} &  \textbf{Teacher / Method ($\downarrow$)} & & Pass@1 & RGR & Pass@1 & RGR & Pass@1 & RGR & Pass@1 & RGR & Pass@1 & RGR & Pass@1 & RGR \\
\midrule[1.5pt]
& Qwen2.5-0.5B (R)  & & 32.20 & -- & 9.78 & -- & 9.93 &  -- & 22.50 & -- & 31.82 &  & 21.25 & -- \\
& Qwen2.5-1.5B (R)  &  & 59.00 & -- & 20.74 & -- & 21.32 & -- & 27.50 & -- & 25.76 & -- & 30.86 & -- \\
\cmidrule(lr){1-15}

\multirow{7}*{\parbox[t]{1.6cm}{\centering Qwen2.5\\Math-7B}}
& \gray Student only & \gray \texttt{Vanilla} & \gray 60.20 & \gray -- & \gray 25.63 & \gray -- & \gray 13.60 & \gray -- & \gray 40.00 & \gray -- & \gray 14.14 & \gray -- & \gray 30.71 & \gray -- \\
&  Student only & \texttt{GRPO} & \textbf{80.20} &  -- &  \textbf{42.37} &  -- & 38.24 &  -- &  57.50 &  -- &  28.28 &  -- &  49.32 &  -- \\
& LIMO & \texttt{SFT} & 70.40 & -- & 34.07 & -- & 30.88 & -- & 47.50 & -- & 17.17 & -- & 40.00 & -- \\

& \blue Qwen2.5-0.5B (R) & \blue \method & \blue 52.00 & \blue 41.25 & \blue 16.30 & \blue 20.01 & \blue 26.84 & \blue 59.73 & \blue 35.00 & \blue 35.71 & \blue 28.79 & \blue 85.59 & \blue 31.79 & \blue 48.46 \\
& \blue Qwen2.5-0.5B (R) & \blue \methodp & \blue 69.20 & \blue 77.08 & \blue 29.33 & \blue 59.99 & \blue 36.40 & \blue 93.50 & \blue 42.50 & \blue 57.14 & \blue 30.81 & \blue 28.53 & \blue 41.65 & \blue 63.25 \\
& \blue Qwen2.5-1.5B (R) & \blue \method & \blue 71.20 & \blue 57.55 & \blue 34.81 & \blue 65.05 & \blue 37.50 & \blue 95.63 & \blue 52.50 & \blue 83.33 & \blue 31.82 & \blue 240.48 & \blue 45.57 & \blue 108.41\\
& \blue Qwen2.5-1.5B (R) & \blue \methodp & \blue \underline{79.00} & \blue 94.34 & \blue \underline{39.26} & \blue 85.62 & \blue \textbf{39.71} & \blue 108.69 & \blue \textbf{62.50} & \blue 116.67 & \blue \textbf{33.33} & \blue 300.40 & \blue \textbf{50.76} & \blue 141.14 \\

\midrule[0.8pt]
\midrule[0.8pt]
& Qwen2.5-1.5B (R)  & & 59.00 & -- & 20.74 & -- & 21.32 & -- & 27.50 & -- & 25.76 & -- & 30.86 & --\\
& Qwen2.5-7B (R) & & 77.80 & -- & 41.78 & -- & 38.97 & -- & 65.00 & -- & 28.79 & -- & 50.47 & --\\
\cmidrule(lr){1-15}

\multirow{7}*{\parbox[t]{1.6cm}{\centering Qwen2.5\\14B}}
& \gray Student only & \gray \texttt{Vanilla} & \gray 62.20 & \gray -- & \gray 25.78 & \gray -- & \gray 28.31 & \gray -- & \gray 35.00 & \gray -- & \gray 28.28 & \gray -- & \gray 35.91 & \gray --\\
& Student only & \texttt{GRPO} &  80.60 &  -- & \textbf{44.74} &  -- & \textbf{47.79} &  -- &  60.00 &  -- &  \textbf{40.40} &  -- &  \textbf{54.71} & --\\
& LIMO & \texttt{SFT} &  75.60 & -- &  \underline{43.85} & -- & 31.25 & -- &  52.50 & -- &  26.26 & -- & 45.89 & --\\
& \blue Qwen2.5-1.5B (R) & \blue \method & \blue 70.20 & \blue 51.85 & \blue 32.89 & \blue 50.63 & \blue 32.72 & \blue 43.07 & \blue 47.50 & \blue 61.54 & \blue 29.80 & \blue 27.60 & \blue 42.62 & \blue 46.94 \\
& \blue Qwen2.5-1.5B (R) & \blue \methodp \; & \blue 76.00 & \blue 78.70 & \blue 36.15 & \blue 64.21 & \blue 38.24 & \blue 63.92 & \blue 45.00 & \blue 53.85 & \blue \underline{39.90} & \blue 96.58 & \blue 47.06 & \blue 71.45 \\
& \blue Qwen2.5-7B (R) & \blue \method & \blue 80.00 & \blue 78.57 & \blue  42.07 & \blue 9.80 & \blue 41.54 & \blue 29.14 & \blue 57.50  & \blue 150.00 & \blue 28.28 & \blue -4.39 & \blue 49.88 & \blue 52.62 \\
& \blue Qwen2.5-7B (R) & \blue \methodp & \blue \textbf{80.80} & \blue 107.14  & \blue 43.56 & \blue 60.14 & \blue \underline{45.22} & \blue 70.86 & \blue \textbf{62.50} & \blue 50.00 & \blue 33.84 & \blue 43.50 & \blue \underline{53.18} & \blue 66.33 \\

\midrule[0.8pt]
\midrule[0.8pt]
& Qwen2.5-1.5B (R) \; & & 59.00 & -- & 20.74 & -- & 21.32 & -- & 27.50 & -- & 25.76 & -- & 30.86 & --\\
& Qwen2.5-7B (R)    & & 77.80 & -- & 41.78 & -- & 38.97 & -- & 65.00 & -- & 28.79 & -- & 50.47 & --\\
& Qwen2.5-14B (R)   & & 80.60 & -- & 44.74 & -- & 47.79 & -- & 60.00 & -- & 40.40 & -- & 54.71 & --\\

\cmidrule(lr){1-15}
\multirow{9}*{\parbox[t]{1.6cm}{\centering Qwen2.5\\32B}}
& \gray Student only & \gray \texttt{Vanilla}\;    & \gray 60.40 & \gray -- & \gray 26.81 & \gray -- & \gray 29.04 & \gray -- & \gray 42.50 & \gray -- & \gray 35.86 & \gray -- & \gray 38.92 & \gray --\\
&  Student only &  \texttt{GRPO} & 82.20 &  -- &  46.96 & -- & \textbf{46.69} &  -- & 72.50 & -- & \textbf{49.49} & -- & 59.57 & --\\
& LIMO & \texttt{SFT} & 75.40 & -- & 47.11 & -- & 30.51 & -- & 62.50 & -- & 32.32 & -- & 49.57 & --\\
& \blue Qwen2.5-1.5B (R) & \blue \method & \blue 70.20 & \blue 48.28 & \blue 31.85 & \blue 42.37 & \blue 37.13 & \blue 62.32 & \blue 52.50 & \blue 55.56 & \blue 35.35 & \blue 40.41 & \blue 45.41 & \blue 49.79 \\
& \blue Qwen2.5-1.5B (R) & \blue \methodp & \blue 76.00 & \blue 73.28 & \blue 37.19 & \blue 62.74 & \blue 40.07 & \blue 73.91 & \blue 62.50 & \blue 77.78 & \blue 38.89 & \blue 55.33 & \blue 50.93 & \blue 68.61\\

& \blue Qwen2.5-7B (R) & \blue \method & \blue  81.60 & \blue 86.36 & \blue 47.41 & \blue 108.69 & \blue 42.28 & \blue 42.88 & \blue 75.00 & \blue 133.33 & \blue 35.86 & \blue 34.15 & \blue 56.43 & \blue 81.08 \\
& \blue Qwen2.5-7B (R) & \blue \methodp & \blue 82.40 & \blue 104.55 & \blue 46.96 & \blue 100.00 & \blue 40.81 & \blue 23.83 & \blue \textbf{77.50} & \blue 166.67 & \blue \underline{44.44} & \blue 75.60 & \blue 58.42 & \blue 94.13 \\

& \blue Qwen2.5-14B (R) & \blue \method & \blue 83.60 & \blue 187.50 & \blue 48.00 & \blue 146.85 & \blue \underline{45.59} & \blue 200.00 & \blue 72.50 & \blue 100.00 & \blue 43.94 & \blue 38.94 & \blue 58.73 & \blue 134.66 \\
& \blue Qwen2.5-14B (R) & \blue \methodp & \blue \textbf{83.80} & \blue 200.00 & \blue \textbf{50.07} & \blue 240.09 & \blue 44.85 & \blue 267.27 & \blue \textbf{77.50} & \blue 140.00 & \blue 42.93 & \blue 27.83 & \blue \textbf{59.83} & \blue 175.04 \\
\bottomrule[1.5pt]
\end{tabular}
\end{adjustbox}
\end{table*}

\begin{figure*}[!h]
    \centering
    \vspace{-2mm}
    \includegraphics[width=\linewidth]{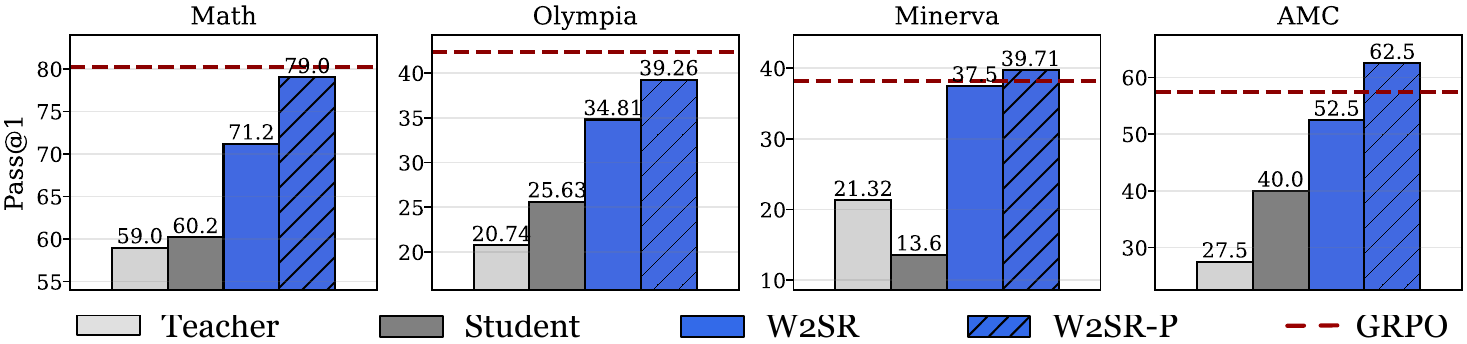}
    \vspace{-6mm}
    \caption{Using \method with \texttt{Qwen2.5-Math-7B} as the strong student and \texttt{Qwen2.5-1.5B-Reasoner} as the weak teacher, the four bars represent: (1) the weak teacher’s standalone performance, (2) the strong student’s standalone performance, (3) the student's performance after \method, and (4) the student's performance after \methodp. More results are provided in~\cref{app:exp}.}
    \vspace{-4mm}
    \label{fig:weak}
\end{figure*}

\subsection{Experimental Setups}
\vspace{-1mm}
Full experimental details are provided in~\cref{app:setup}, all prompts are included in~\cref{app:prompt}.
\vspace{-5mm}
\paragraph{Dataset}
We use the MATH dataset~\cite{hendrycks2measuring} with difficulty levels 3–5 as the training data, following~\citep{zeng2025simplerl}.
This dataset encompasses seven math topics such as advanced calculus, geometry, and linear algebra. 

\paragraph{Models} 
The student models include three parameter sizes from the Qwen-2.5 family~\cite{qwen2.5}: \texttt{Qwen2.5-Math-7B}, \texttt{Qwen2.5-14B}, and \texttt{Qwen2.5-32B}.
The teacher models are from \texttt{Qwen-2.5} (0.5B, 1.5B, 7B, 14B, 32B), for each we evaluate both the base version and its reinforcement learning fine-tuned counterpart.
All reinforcement learning fine-tuned models are sourced from SimpleRL-Zoo~\cite{zeng2025simplerl}, where GRPO~\cite{shao2024deepseekmath} is the training algorithm.

\paragraph{Training}
Our knowledge distillation framework involves two stages: (1) data distillation with teacher models using vLLM~\cite{kwon2023efficient} (greedy decoding, temperature 0, top-p 1, max length 4096 tokens, one sample per input); and (2) supervised fine-tuning (SFT) of student models using LLaMA-Factory~\cite{zheng2024llamafactory} with batch size 128, learning rate $10^{-5}$, for 5 epochs.

\paragraph{Evaluation}
We evaluate on five reasoning benchmarks, including three standard math datasets, MATH500~\cite{hendrycks2measuring}, OlympiaBench~\cite{he2024olympiadbench}, MinervaMath~\cite{lewkowycz2022minerva}, plus the competition-level AMC2023~\cite{amc23dataset} and the non-mathematical GPQA~\cite{rein2024gpqa}. Evaluation uses the codebase from~\cite{li2025limr}, with sampling temperature 0.6, top-p 0.95, 1 sample per input, and max generation length 32,768 tokens.

\section{Experiments}
\label{sec:exp}

The remainder of paper analyzes weak-to-strong reasoning paradigm from three key perspectives:
\textbf{(RQ1)} Can weak supervision incentivize reasoning in stronger models?
\textbf{(RQ2)} What aspects of teacher supervision are most critical for incentivizing reasoning?
\textbf{(RQ3)} What practical benefits and broader impacts arise from weak-to-strong reasoning?

\subsection{RQ1: Can Weak Supervision Incentivize Reasoning in Stronger Models?}

In this section, we examine whether high-capacity student models can improve their reasoning capabilities when trained on reasoning trajectories produced by significantly smaller and less capable teacher models. We also assess how effectively such weak-to-strong supervision can narrow the performance gap relative to models fine-tuned with RL).
As illustrated in~\cref{tab:weak} and~\cref{fig:weak}, we evaluate three student models ranging from 7B to 32B parameters, each trained using supervision from multiple weaker teachers, whose sizes range from 0.5B to 14B parameters.

Based on the results in~\cref{tab:weak}, we evaluate the effectiveness of weak-to-strong reasoning for each strong student model with respect to its corresponding weak teacher(s), where the teacher possesses reasoning capabilities (denoted as ``R'').
For each teacher-student pair, we report both Pass@1 and Relative Gain in Reasoning (RGR) metrics under both \method and \methodp.
To ensure a comprehensive evaluation, we additionally assess model performance on LIMO~\cite{ye2025limo}, a benchmark constructed from  high-quality CoT data.

As shown in~\cref{fig:weak}, we analyze the training of the \texttt{Qwen2.5-Math-7B} student with the \texttt{Qwen2.5-1.5B-Reasoner} as the teacher, comparing five settings: (1) the standalone weak teacher, (2) the standalone student, (3) the student trained via \method, (4) the student trained via \methodp, and (5) the student trained via RL.
These comparisons reveal a striking effect, where combining a weak teacher with a strong student results in performance far exceeding that of either model alone.
For instance, on MATH and AMC, the original student achieves Pass@1 of 60.20\% and 40.00\%, while the teacher scores 59.00\% and 27.50\%, respectively, both considerably lower.
However, after \methodp training, despite the teacher being 4.7× smaller and performing 1.99\% and 31.25\% relatively worse than the student, it enables the student to reach Pass@1 of 79.00\% and 62.50\%, corresponding to relative improvements of 31.22\% and 56.25\%, reflected in RGR of 94.34\% and 116.67\%.
Notably, on AMC, this improvement even surpasses the performance ceiling established by RL training.

Our findings reveal that teachers with structured reasoning abilities, regardless of model scale or final-answer accuracy, can effectively elicit and enhance reasoning capabilities in more powerful student models. This knowledge transfer enables students to not only surpass their teachers' performance but also exceed results from direct reinforcement learning on the student models themselves, aligning with the insight shown in \underline{Takeaway 1}.

\begin{AIbox}{Takeaway 1: Weak Yet Incentivizable}
\textbf{Weak Supervision Can Incentivize Reasoning in Stronger Models}:
Reasoning trajectories from weaker teachers can effectively incentivize reasoning in stronger students and reach comparable/superior performance compared to expensive RL methods.
\end{AIbox}

\subsection{RQ2: Key Aspects of Teacher Supervision for Incentivizing Reasoning?}

\begin{figure*}[!h]
    \centering
    \includegraphics[width=\linewidth]{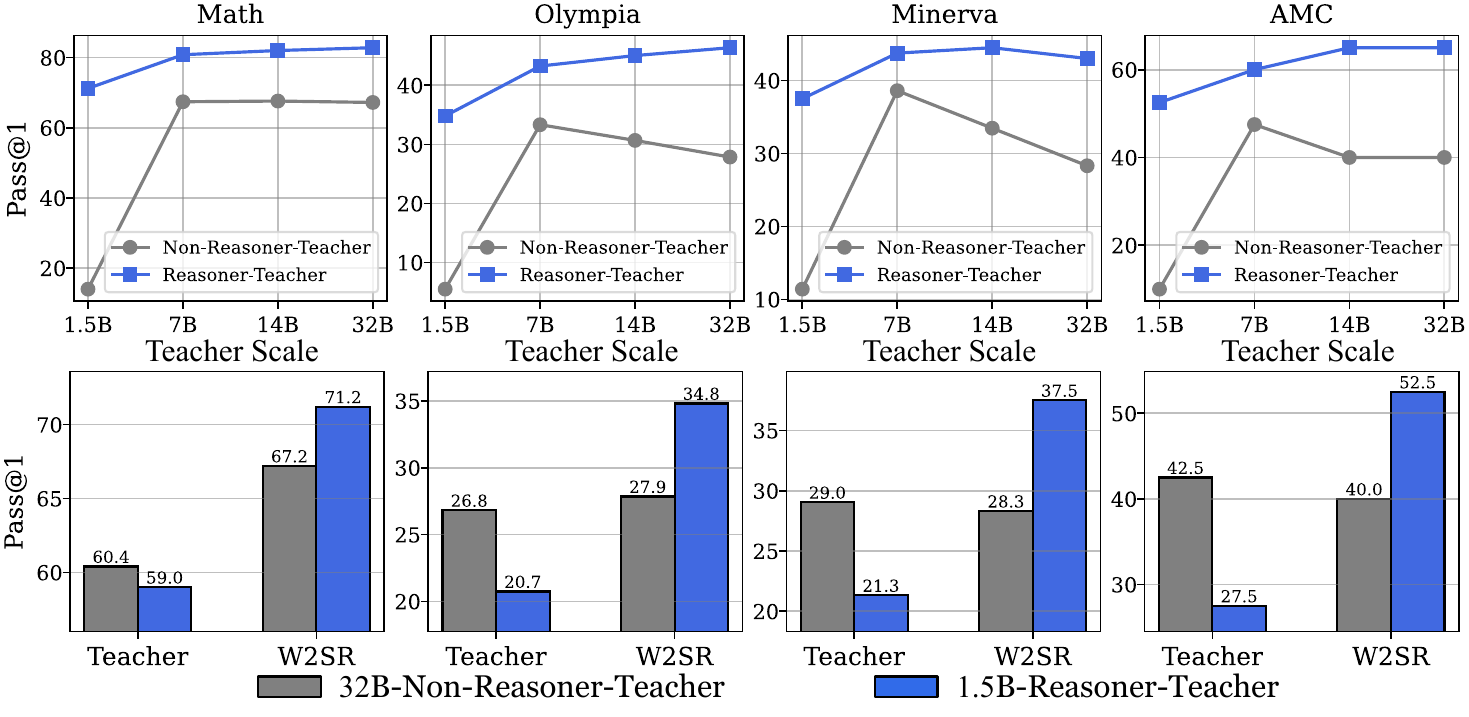}
    \caption{Comparison of Weak-to-Strong Reasoning (\method) between Reasoner and Non-Reasoner Teachers using \texttt{Qwen2.5-Math-7B} as the student. \textbf{Upper}: The x-axis represents teacher model scale, with two lines indicating the \method student performance trained with Reasoner vs. Non-Reasoner teachers of corresponding scales. \textbf{Lower}: The first two bars show initial performance of \texttt{Qwen2.5-1.5B-Reasoner} and \texttt{Qwen2.5-32B-Non-Reasoner} teacher, while the last two bars show the resulting \method student performance. More results are in~\cref{app:exp}.}
    \label{fig:non}
\end{figure*}

We ask what makes weak supervision effective for transferring reasoning to stronger students. Specifically:  
(1) Which attributes, reasoning ability, model scale, or task accuracy, are most critical for supervision?  
(2) Must supervision trajectories yield correct answers, or can incorrect ones still be useful?

\begin{table*}[t]
\centering
\caption{Pass@1 Performance for three variants of Weak-to-Strong Reasoning: (1) All: Unfiltered reasoning trajectories (\method, symbolically denoted as \rnw), (2) Correct Only: Reasoning trajectories filtered by correct answers (\methodp, symbolically denoted as \onlyr), and (3) Incorrect Only: Reasoning trajectories filtered by incorrect answers (\methodn, symbolically denoted as \onlyw).}
\label{tab:wrong}
\begin{adjustbox}{width=0.88\textwidth}
\begin{tabular}{cclcccccc}
\toprule[1pt]
\textbf{Student} & \multicolumn{2}{c}{\textbf{Teacher}}   &
\textbf{Math} & \textbf{Olympic} & \textbf{Minera} & \textbf{AMC} & \textbf{GPQA} & \textbf{Average}\\
\midrule
\multirow{4}{*}{\parbox[t]{1.5cm}{\centering Qwen2.5-Math-7B}} &
Student only & Vanilla & 60.20 & 25.63 & 13.60 & 40.00 &  14.14 &  30.71\\
\cmidrule(lr){2-9}
& \multirow{3}{*}{\parbox[t]{1.5cm}{\centering Qwen2.5-1.5B (R)}} &\method\;(\rnw\,)& 71.20 & 34.81 & 37.50 & 52.50 & 31.82 & 45.57\\
& &\methodp\;(\onlyr)& 79.00 &  39.26 & 39.71 & 62.50 & 33.33 & 50.76\\
\cmidrule(lr){3-9}
& &\methodn\;(\onlyw\,)& \blue 70.20 & \blue 32.44 & \blue 37.87 & \blue 42.50 & \blue 35.86 & \blue  43.77\\
\midrule

\multirow{4}{*}{\parbox[t]{1.5cm}{\centering Qwen2.5\\14B}} &
Student only & Vanilla &  62.20 & 25.78 & 28.31 & 35.00 & 28.28 & 35.91 \\
\cmidrule(lr){2-9}
& \multirow{3}{*}{\parbox[t]{1.5cm}{\centering Qwen2.5-7B (R)}} 
&\method\;(\rnw\,)& 78.80 & 42.96 & 40.07 & 60.00 & 28.28 & 50.02 \\
& &\methodp\;(\onlyr)&  80.80 &  43.56 & 45.22 & 62.50  & 33.84 & 53.18 \\
\cmidrule(lr){3-9}
& &\methodn\;(\onlyw\,)& \blue 78.20 & \blue 41.04 & \blue 40.81 & \blue 55.00 & \blue 36.36 & \blue 50.28 \\
\midrule

\multirow{4}{*}{\parbox[t]{1.5cm}{\centering Qwen2.5\\32B}} &
Student only & Vanilla & 60.40 & 26.81 & 29.04 & 42.50 & 35.86 & 38.92 \\
\cmidrule(lr){2-9}
& \multirow{3}{*}{\parbox[t]{1.5cm}{\centering Qwen2.5-14B (R)}} 
&\method\;(\rnw\,)& 83.60 & 48.00 & 45.59 & 72.50 & 43.94 & 58.73 \\
& &\methodp\;(\onlyr)& 83.80 & 50.07 & 44.85 & 77.50 & 42.93 & 59.83 \\
\cmidrule(lr){3-9}
& &\methodn\;(\onlyw\,)& \blue 83.20  & \blue 47.41  & \blue 46.69 & \blue 62.50 & \blue 39.90 & \blue 55.94 \\
\bottomrule[1pt]
\end{tabular}
\end{adjustbox}
\vspace{-4mm}
\end{table*}
\begin{figure*}[!t]
    \centering
    \vspace{-3mm}
    \includegraphics[width=0.95\linewidth]{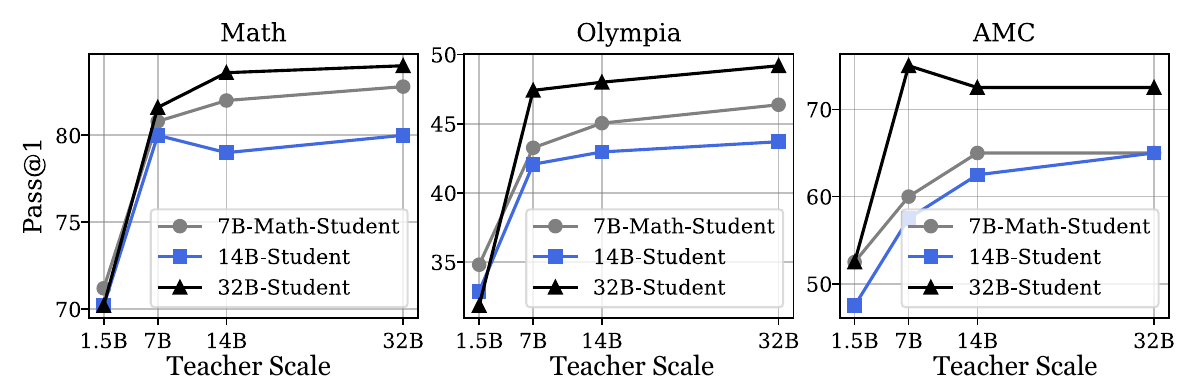}
    \caption{Diminishing marginal returns from increasing reasoner-teacher scale on student performance. The x-axis represents model sizes/scale of the reasoner teacher, increasing from 1.5B, 7B, 14B to 32. The y-axis shows the Pass@1 performance of students trained under supervision from the corresponding reasoner-teacher. The three lines correspond to student performance of different sizes (7B-Math, 14B, and 32B). The results show that as teacher strength increases, student performance improves initially but eventually plateaus, indicating limited additional benefit from further increasing teacher scale.}
    \label{fig:limit}
    \vspace{-3mm}
\end{figure*}
\subsubsection{What Makes a Good Teacher for Incentivizing Reasoning?}

This section provides empirical evidence that the capability of a teacher model’s explicit inference-time scaling plays a more critical role
than parameter scaling or overall task accuracy.

\paragraph{(1) Reasoning Ability vs. Model Scale.}
As shown in the upper part of~\cref{fig:non}, across four benchmarks, students trained with Reasoner teachers consistently outperform those trained with Non-Reasoner teachers, regardless of the teacher’s parameter count. Notably, increasing the size of Non-Reasoner teachers from 1.5B to 32B yields no meaningful improvement in student performance. In most cases, students trained with the smallest Reasoner teacher (1.5B) surpass those trained with even the largest Non-Reasoner teacher (32B), despite the latter being 21× larger in model size. These results indicate that scaling model size alone is insufficient for effective supervision, what matters more is the presence of explicit reasoning trajectories in the teacher’s outputs.

\paragraph{(2) Reasoning Ability vs. Task Performance.}
The lower part of~\cref{fig:non} highlights a direct comparison between two extreme cases: the small reasoner teacher (\texttt{Qwen2.5-1.5B-Reasoner}) and the large Non-Reasoner teacher (\texttt{Qwen2.5-32B-Non-Reasoner}). While the 1.5B Reasoner teacher performs worse in isolation, e.g., on AMC it scores 27.5\%, compared to 32B's 42.5\%, it still leads to significantly better student models across all four benchmarks. For example, the student trained with the 1.5B Reasoner teacher achieves a Math Pass@1 of 71.2\%, compared to 67.2\% for the one trained with the 32B Non-Reasoner as teacher. 
The effect is even more pronounced on Minerva and AMC, where using the 32B Non-Reasoner teacher actually results in student performance worse than the teacher by 2.41\% and 5.88\% relatively, the 1.5B Reasoner improves it by 76.06\% and 90.91\%.

These findings underscore that reasoning supervision, enabled by inference-time scaling through explicit CoT traces, is fundamentally more effective than simply increasing model size. Although larger models may excel at end-task metrics, they often lack the structured and interpretable inference procedures necessary to guide student learning. In contrast, weak reasoner, despite weaker standalone accuracy, consistently produce stronger \method students by providing step-by-step, traceable reasoning. Overall, our results affirm that inference scaling is more critical than parameter scaling in fostering reasoning capabilities in student models, echoing the insight summarized in \underline{Takeaway 2}.

\begin{AIbox}{Takeaway 2: Reasoning > Size + Acc}
\textbf{Reasoning Capability Matters More Than Model Size or Accuracy}:
A teacher’s explicit reasoning capability plays a more critical role than its parameter scale or overall task accuracy for effectively incentivizing student reasoning ability.
\end{AIbox}

\subsubsection{Is Correctness a Necessary Condition for Useful Reasoning Supervision?}
We investigate whether effective reasoning supervision requires correct final answers or if reasoning traces with incorrect outcomes can still provide useful learning signals in weak-to-strong training.

We conduct experiments on three different types of weak teacher trajectories (1) All: Unfiltered reasoning trajectories (\method), (2) Correct Only: Reasoning trajectories filtered by correct answers (\methodp), and (3) Incorrect Only: Reasoning trajectories filtered by incorrect answers (\methodn). 

As shown in \cref{tab:wrong}, student models trained with \methodn consistently outperform vanilla students (trained without teacher supervision) across all benchmarks and model scales.
For example, the \texttt{Qwen2.5-Math-7B} student distilled with incorrect traces from the \texttt{Qwen2.5-1.5B-Reasoner} teacher improves its average Pass@1 from 30.71\% to 43.77\%, even surpassing \method. Similarly, the \texttt{Qwen2.5-14B} student distilled from the \texttt{Qwen2.5-7B-Reasoner} teacher achieves Pass@1 scores of 50.02\% with All, 53.53\% with Correct Only, and still a strong 50.28\% with Incorrect Only.
These results demonstrate that intermediate reasoning steps can remain pedagogically valuable even when the final answers are incorrect. 

Our findings support \underline{Takeaway 3}, showing that the correctness of final answer alone is an insufficient criterion for high-quality supervision. Instead, structurally sound reasoning traces, even when yielding incorrect final answers, can effectively activate reasoning capabilities in student models.

\begin{AIbox}{Takeaway 3: Wrong Yet Helpful}
\textbf{Incorrect Answers Can Still Teach Correct Reasoning}:
Incorrect reasoning trajectories can still effectively incentivize the reasoning capability of the student model, showing that correctness is not essential for effective supervision.
\end{AIbox}

\subsection{RQ3: What Practical Benefits Arise from Weak-to-Strong Reasoning?}

In this section, we investigate the benefits of our \method framework through the perspectives of effectiveness, efficiency and further analysis of cognitive behaviors and inference-time scaling.

\subsubsection{Effectiveness: Beyond Strong Supervision and Base Model Constraints}

Current reasoning methods, supervised fine-tuning (SFT) with strong supervision and reinforcement learning (RL), face inherent limitations. As models improve, high-quality strong supervision becomes scarce, limiting further gains.  
Moreover, RL heavily depends on the base model’s existing abilities, with the Pass@K gap between base and RL-tuned models shrinking as K increases~\cite{yue2025does}, suggesting RL mainly amplifies known capabilities rather than enabling novel reasoning.  

In contrast, our \method employs a weak-to-strong paradigm that leverages weak supervision to elicit and enhance reasoning ability without relying on scarce expert labels, enabling acquisition of new knowledge beyond the base model’s priors and overcoming limitations of both SFT and RL.

\subsubsection{Efficiency: Weaker Teachers are Sufficient for Incentivization}

\begin{figure}[!t]
\centering
\includegraphics[width=\linewidth]{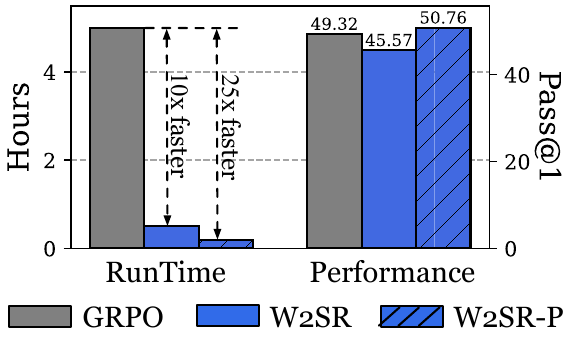}
\caption{Comparison of efficiency (left bars) and performance (right bars) among GRPO, \method, and \methodp, evaluated with the \texttt{Qwen2.5-Math-7B} student and the \texttt{Qwen2.5-1.5B-Reasoner} teacher. \methodp achieves a 25× improvement in training efficiency and a 2.92\% gain in performance. Further results are in~\cref{app:exp}.}
\label{table:cost} 
\end{figure}

Compared to SFT relying on large expert models and RL, which is unstable and costly, our \method paradigm offers a more efficient alternative with competitive performance.
As shown in~\cref{fig:limit}, increasing teacher size yields diminishing returns: on Math with a \texttt{Qwen2.5-32B} student, scaling the teacher from 1.5B to 7B improves performance by 11.4\%, but scaling further to 32B adds only 0.4\%.
In~\cref{table:cost}, comparing GRPO, \method, and \methodp with a \texttt{Qwen2.5-Math-7B} student and 1.5B teacher, \methodp cuts training time by 25× and improves performance by 2.92\% over GRPO.
These results show weaker teachers suffice for effective reasoning incentivization, offering a better efficiency–performance trade-off. We summarize the above important findings in \underline{Takeaway 4}.

\begin{AIbox}{Takeaway 4: Weak Is Sufficient}
\textbf{Weaker Teachers Offer Better Efficiency–Performance Trade-off}:
Scaling teacher strength provides diminishing returns; notably, weak teachers often suffice for effectively incentivizing reasoning, significantly reducing computational cost.
\end{AIbox}

\subsubsection{Further Analysis}

\paragraph{Cognitive Behaviors}

\begin{table}[!t]
\centering

\captionof{table}{Cognitive Behavior of  of reinforcement learning and \method. For detailed case studies, please refer to \cref{app:case}.}

\resizebox{0.49\textwidth}{!}{
\begin{tabular}{cccc}
\toprule
\textbf{Model} & \begin{tabular}[c]{@{}c@{}}\textbf{Structured}\\  \textbf{Reasoning}\end{tabular} & \begin{tabular}[c]{@{}c@{}}\textbf{Correct}\\ \textbf{Answer}\end{tabular} & \textbf{Length}  \\
\midrule
Weak Teacher& \ding{51} & \ding{55}& 1038 \\
Base Student & \ding{55} & \ding{55} & 307 \\
RL Student & \ding{51} & \ding{51} & 1038 \\
\method Student & \ding{51} & \ding{51} & 984 \\
\bottomrule
\end{tabular}}
\label{tab:cog} 
\end{table}
Our case study highlights how \method can induce advanced cognitive behaviors. As shown in~\cref{tab:cog}, with full outputs provided in~\cref{app:case}, the weak teacher produces a structurally valid but numerically incorrect solution. The \method student retains this structured reasoning while correcting the computational errors, ultimately arriving at the correct answer. Its reasoning depth increases significantly, from 307 tokens (base student) to 984 tokens, approaching the length and quality of the RL-trained student (1038 tokens). These findings suggest that \method effectively leverages flawed yet structured supervision to elicit strong reasoning capabilities.

\vspace{2mm}
\paragraph{Inference-time Scaling}
We investigate whether \method induces a notable increase in response length during inference time, similar to RL. As shown in~\cref{app:length}, generation length steadily increases in training, reflecting long CoT and signaling the emergence of inference-time scaling.

\section{Conclusion}
In this work, we introduce \method, a simple yet effective paradigm  for incentivizing the reasoning abilities of large language models (LLMs) through supervision from significantly weaker teachers. Instead of relying on expensive reinforcement learning or expert-quality chain-of-thought demonstrations, \method leverages structured, albeit imperfect, reasoning trajectories to elicit strong reasoning behaviors. 
Our experiments across multiple benchmarks show that student models trained with \method consistently outperform their weak teachers and even rival RL-based approaches, while requiring drastically less computation. 
These findings position weak-to-strong reasoning as a scalable and cost-effective alternative for incentivizing strong reasoning capabilities in LLMs.
This paradigm opens several directions for future research, including improving the selection and aggregation of weak teachers, extending to multi-modal or tool-augmented settings, and developing adaptive techniques for filtering or refining imperfect reasoning trajectories. 
We hope this work encourages further exploration into scalable and cost-efficient approaches for reasoning incentivization in large language models, paving the way toward scalable oversight that enables strong reasoning capabilities to be widely attainable.

\clearpage

\bibliography{main}

\newpage
\appendix

\section{Appendix Summary}

\begin{itemize}[leftmargin=1em]
\item \textbf{Experimental Details} (\cref{app:setup}): 
\begin{itemize}
    \item Datasets (\cref{app:dataset})
    \item Evaluation metrics (\cref{app:evaluation})
    \item Models (\cref{app:model})
    \item Implementation (\cref{app:implementation})
    \item Compute resources (\cref{app:resources})
    
\end{itemize}
\item \textbf{Additional Experiments} (\cref{app:exp}):
\begin{itemize}
    \item Inference Scaling (\cref{app:length})
    \item Additional Results for Takeaway 1 \\ (\cref{app:take1})
    \item Additional Results for Takeaway 2 \\ (\cref{app:take2})
    \item Additional Results for Takeaway 4 \\ (\cref{app:take3})
\end{itemize}
\item \textbf{Limitations and Broader Impacts} \\ (\cref{app:impacts})
\item \textbf{More Related Works} (\cref{app:related}).
\item \textbf{Insights and Future directions} (\cref{app:future})
\item \textbf{Prompt Details} (\cref{app:prompt}).
\item \textbf{Case Studies} (\cref{app:case}).
\end{itemize}

\section{Experimental Details}
\label{app:setup}

\subsection{The Details of Datasets} 
\label{app:dataset}

\paragraph{MATH}~\cite{lightman2023lets} contains 500 high-school math competition problems, which are of higher difficulty and complexity, requiring more in-depth mathematical reasoning ability. 

\paragraph{OlympiadBench}~\cite{he2024olympiadbench} is a bilingual, multimodal scientific benchmark at the Olympiad level, comprising 8,476 problems from advanced mathematics and physics competitions, including the Chinese college entrance examination. Each problem is accompanied by expert-level annotations that provide step-by-step reasoning. The standard test benchmark contains 675 problems.

\paragraph{Minerva}~\cite{lewkowycz2022solving} is a dataset of undergraduate-level questions in science and mathematics from MIT’s OpenCourseWare.

\paragraph{AMC} is a 2023 middle school math competition covering arithmetic, algebra, geometry, and more. It consists of 40 questions.

\paragraph{GPQA}~\cite{rein2024gpqa} is a challenging dataset of 448 multiple-choice questions written by domain experts in biology, physics, and chemistry. We use GPQA Diamond, the highest quality subset consisting of 198 questions, including only questions that both experts answer correctly and the majority of non-experts answer incorrectly.

\subsection{The Details of Evaluation Metrics} 
\label{app:evaluation}
\textbf{Pass@k} Given a total of $N$ problems, for each problem $i$, we sample $k$ responses. Let $C_i$ be the number of correct responses among the $k$ samples for problem $i$, and let $n_i$ be the total number of possible distinct completions considered for that problem. Then the Pass\@k accuracy is defined as:

\begin{equation}
\mathrm{Pass@k} = \frac{1}{N} \sum_{i=1}^{N} \left[1 - \frac{\binom{n_i - C_i}{k}}{\binom{n_i}{k}}\right]
\end{equation}

\subsection{The Details of Models}
\label{app:model}

\paragraph{Qwen-2.5}~\cite{yang2024qwen2} is the next-generation open-source large language model series developed by the Qwen team. It includes models ranging from 0.5B to 72B parameters, trained on high-quality multilingual corpora. Qwen2.5 models demonstrate strong performance across a wide range of benchmarks, such as MMLU, GSM8K, and HumanEval, and are particularly effective in Chinese language understanding and generation. We utilize Qwen2.5-Math-7B, Qwen2.5-14B, and Qwen2.5-32B. 

\paragraph{SimpleRL-Zoo}~\cite{zeng2025simplerl} is a model family fine-tuned using the zero RL training algorithm on MATH training datasets, spanning a diverse range of model series and sizes, including Mistral-7B, Mistral-24B, LLaMA3-8B, DeepSeek-Math-7B, Qwen2.5-0.5B/1.5B/7B/14B/32B, and Qwen2.5-Math-7B.

\subsection{The Details of Implementation}
\label{app:implementation}

We conduct model training using LLaMA-Factory~\cite{zheng2024llamafactory}, applying full-parameter fine-tuning to all student models, ranging from 7B to 32B. The hyperparameters used are summarized in~\cref{tab:hyper}.
Although the global batch size is fixed at 128 for all models, the per-device batch size, gradient accumulation steps, and number of GPUs vary across model scales. Specifically, for the 7B model, we use 4 GPUs with a per-device batch size of 2 and a gradient accumulation step of 16 (4GPU * 2PBS * 16GA); for the 14B model, we use 4 GPUs with a per-device batch size of 1 and a gradient accumulation step of 32 (4GPU * 1PBS * 32GA); and for the 32B model, we adopt 8 GPUs with a per-device batch size of 1 and a gradient accumulation step of 16 (8GPU * 1PBS * 16GA).

\begin{table}[!h]
\centering
\vspace{-3mm}
\caption{Hyperparameters used for full-parameter fine-tuning.}
\begin{tabular}{ll}
\toprule
\textbf{Hyper-parameter} & \textbf{Value} \\ \midrule
Learning Rate & $1 \times 10^{-5}$ \\
Number of Epochs & $10$ \\
Global Batch Size & $128$ \\
Optimizer & \texttt{Adamw} \\
Learning Rate Scheduler & \texttt{cosine} \\
Max Sequence Length  & $4096$ \\ \bottomrule
\end{tabular}
\label{tab:hyper}
\vspace{-5mm}
\end{table}

\subsection{The Details of Compute Resources}
\label{app:resources}
All the training experiments in this paper were conducted on 4 $\times$ NVIDIA A100 (80G) GPUs.

\section{Additional Experiments}
\label{app:exp}

\subsection{Inference-time Scaling}
\label{app:length}
This section examines the impact of our method on inference-time behavior~\cite{yuan2025inference,yuan2024fact}, with a particular focus on response length.
We investigate whether \method leads to an increase in generation length during inference, similar to trends observed with reinforcement learning.
As shown in~\cref{tab:length}, using \texttt{Qwen2.5-Math-7B}, \texttt{Qwen2.5-14B} and \texttt{Qwen2.5-32B} as student and \texttt{Qwen2.5-1.5B-Reasoner} as teacher on  AMC, the generation length increases steadily throughout training, indicative of longer CoT and suggesting the emergence of inference-time scaling effects.

\begin{table*}[!h]
\centering
\caption{Average Response Length.}
\vspace{-2mm}
\begin{adjustbox}{width=0.85\textwidth}
\setlength{\tabcolsep}{2mm}
\begin{tabular}{llccc}
\toprule
\textbf{Teacher} & \textbf{Method} &\textbf{Qwen2.5-Math-7B} & \textbf{Qwen2.5-14B} & \textbf{Qwen2.5-32B}\\ \midrule
Student only & Vanilla & 882.78 & 643.90  & 600.20\\
Student only & \texttt{GRPO} & 1122.27 & 1003.05 & 1120.56  \\
\texttt{Qwen2.5-1.5B(R)} &\method & 1545.92 & 1559.47 & 1373.55 \\
\bottomrule
\end{tabular}
\end{adjustbox}
\vspace{-4mm}
\label{tab:length}
\end{table*}

\subsection{Additional Results for Takeaway 1}
\label{app:take1}
\begin{figure*}[!h]
    \centering
    \includegraphics[width=\linewidth]{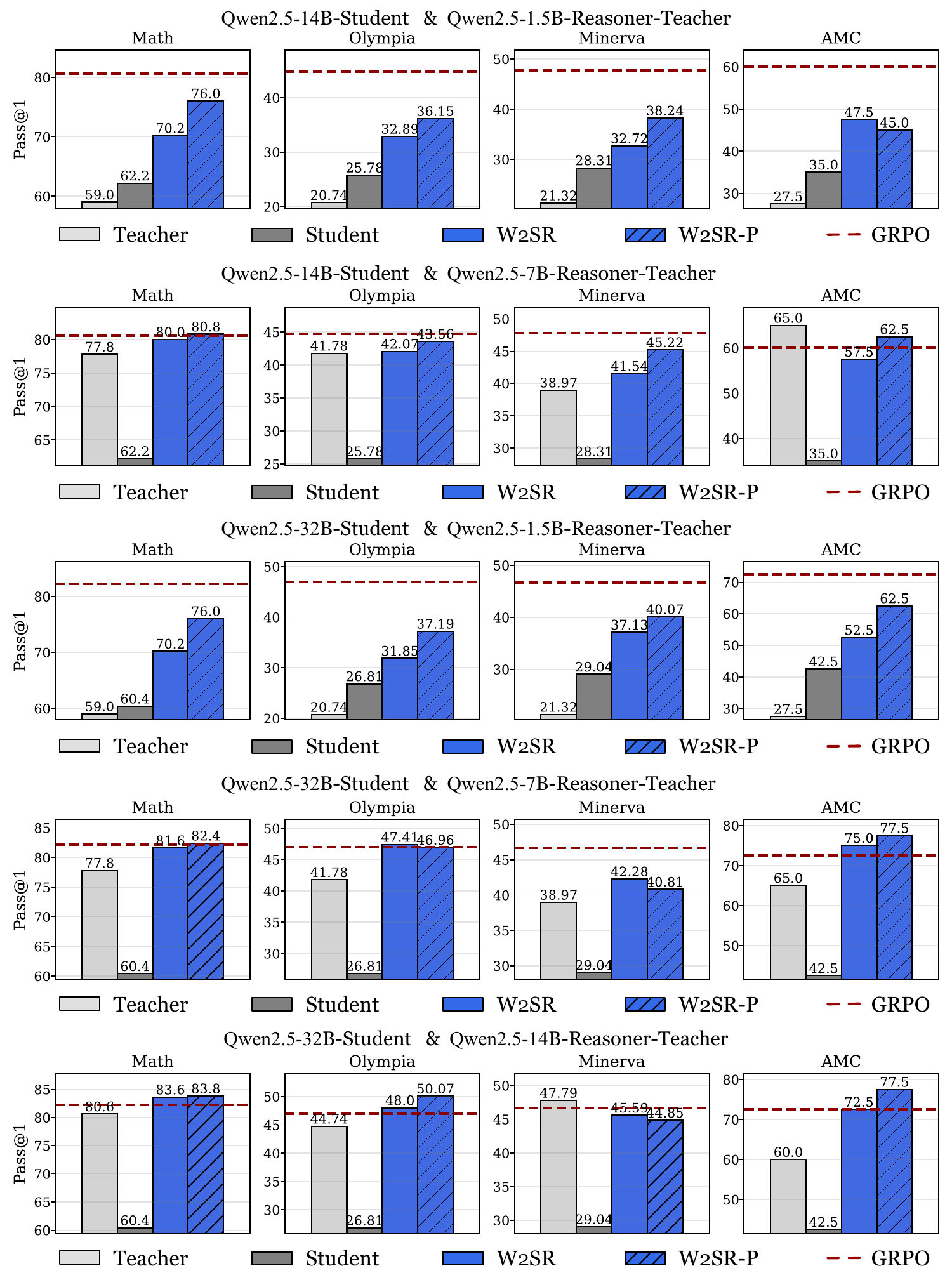}

    \caption{From top to bottom: Using (1) \texttt{Qwen2.5-14B} as the strong student and \texttt{Qwen2.5-1.5B-Reasoner} as the weak teacher. (2) \texttt{Qwen2.5-14B} as the strong student and \texttt{Qwen2.5-7B-Reasoner} as the weak teacher. (3) \texttt{Qwen2.5-32B} as the strong student and \texttt{Qwen2.5-1.5B-Reasoner} as the weak teacher. (4) \texttt{Qwen2.5-32B} as the strong student and \texttt{Qwen2.5-7B-Reasoner} as the weak teacher. (5) \texttt{Qwen2.5-32B} as strong student and \texttt{Qwen2.5-14B-Reasoner} as weak teacher. Four bars represent: teacher’s standalone performance, student’s standalone performance, student performance after \method, student performance after \methodp.}
    \label{fig:weak-app}
\end{figure*}

This section presents supplementary results supporting Takeaway 1: weak supervision can incentivize reasoning in stronger models. Specifically, we show that reasoning trajectories derived from weaker teacher models can effectively encourage stronger student models to engage in reasoning, achieving performance comparable to or exceeding that of more costly reinforcement learning approaches. Figure~\ref{fig:weak-app} presents five additional configurations, each illustrating a distinct teacher–student pairing. Across all settings, student models demonstrate substantial gains when supervised by weaker reasoners. Notably, the performance improvements achieved by our proposed method (\method) and its enhanced variant (\methodp) consistently surpass those of baseline standalone students, and in some cases, even outperform students trained with expensive reinforcement learning.

\subsection{Additional Results for Takeaway 2}
\label{app:take2}
\begin{figure*}[!h]
    \centering
    \includegraphics[width=\linewidth]{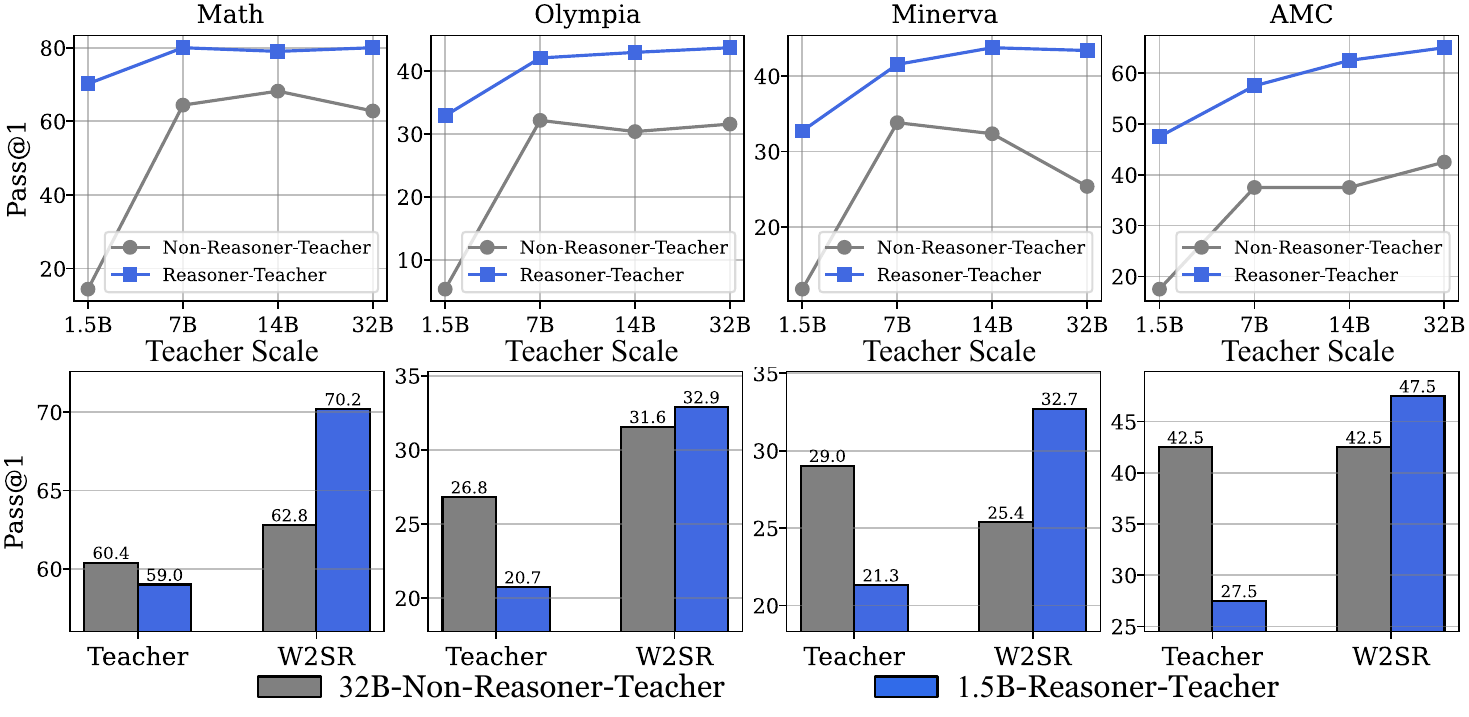}
    \caption{Comparison of Weak-to-Strong Reasoning (\method) between Reasoner and Non-Reasoner Teachers using \texttt{Qwen2.5-14B} as the student. \textbf{Upper}: The x-axis represents teacher model scale, with two lines indicating the \method student performance trained with Reasoner vs. Non-Reasoner teachers of corresponding scales. \textbf{Lower}: first two bars show initial performance of \texttt{Qwen2.5-1.5B-Reasoner} and \texttt{Qwen2.5-32B-Non-Reasoner} teacher, last two bars show the resulting \method student performance.}
    \label{fig:non-app}
\end{figure*}

This section provides supplementary results supporting Takeaway 2: reasoning capability is more important than model size or overall accuracy. We observe that a teacher’s ability to perform explicit reasoning plays a more critical role than its parameter count or average task accuracy in fostering reasoning skills in student models. Figure~\ref{fig:non-app} presents additional configurations where \texttt{Qwen2.5-14B} serves as the student model. Across these settings, students consistently benefit more from teachers with explicit reasoning abilities, even when those teachers are smaller or less accurate, compared to larger teachers lacking such abilities.

\subsection{Additional Results for Takeaway 4}
\label{app:take3}
\begin{figure*}[!h]
    \centering
    \includegraphics[width=\linewidth]{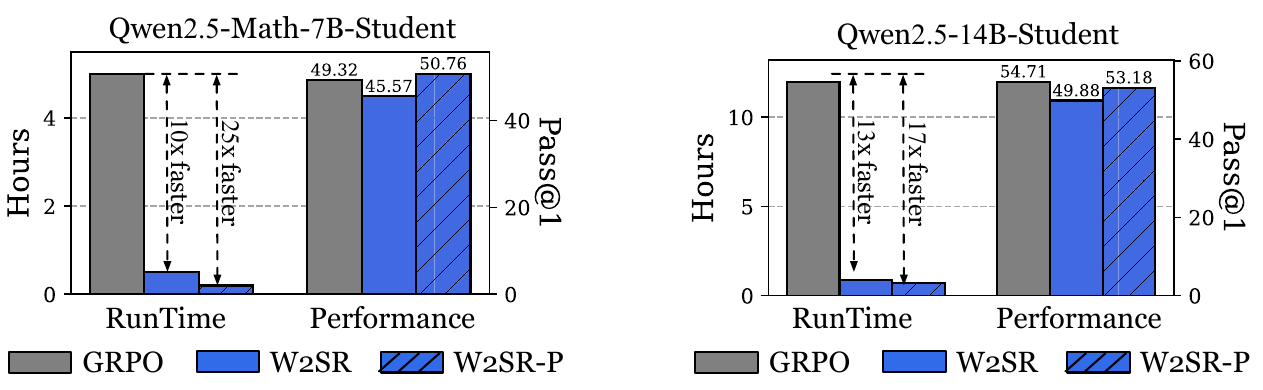}
    \caption{Comparison of efficiency and performance among GRPO, \method, and \methodp. Left: \texttt{Qwen2.5-Math-7B} as the student and \texttt{Qwen2.5-1.5B-Reasoner} as the teacher. Right: \texttt{Qwen2.5-14B} as the student and \texttt{Qwen2.5-7B-Reasoner} as the teacher. For each, the left three bars represent the time cost of GRPO, \method, and \methodp respectively, the right three bars show the corresponding performance outcomes.}
    \label{fig:cost-app}
\end{figure*}

This section provides supplementary results supporting Takeaway 4, which suggests that weaker teachers offer a more favorable efficiency–performance trade-off. As shown in Figure~\ref{fig:cost-app}, weaker teachers can achieve performance comparable to reinforcement learning-based methods while substantially reducing training costs. This demonstrates their practicality in resource-constrained settings.

\section{Limitations and Broader Impacts}
\label{app:impacts}
This work focuses on mathematical reasoning, and the effectiveness of \method in broader domains such as commonsense reasoning, scientific QA, or legal analysis remains to be verified. The approach assumes that weak teacher models can produce structured reasoning traces, which may not hold in low-resource or complex domains. Additionally, while imperfect reasoning trajectories are often helpful, unfiltered or low-quality supervision may introduce noise and reduce robustness. Current method also relies on SFT and may benefit from more adaptive training strategies in future work.

\method provides a practical and efficient approach to improving reasoning abilities in language models without relying on expensive data collection or RL. This can enhance access to powerful models in academic and low-resource settings. However, enabling stronger reasoning capabilities may also increase the risk of models producing convincing but flawed outputs, especially when trained on imperfect supervision. Applications in high-risk domains should be accompanied by safeguards such as rigorous evaluation, interpretability tools, and human oversight to mitigate potential harms.

\section{Insights and Future Directions}
\label{app:future}
Our work shows that structured but imperfect reasoning traces from significantly weaker models can effectively incentivize reasoning in stronger LLMs, rivaling reinforcement learning at a fraction of the cost. Notably, reasoning structure, rather than teacher size or accuracy, plays a pivotal role. Incorrect traces remain pedagogically valuable, and increasing teacher scale yields diminishing returns.

Future work includes adaptive filtering of reasoning trajectories to prioritize useful supervision, leveraging ensembles of weak teachers to provide richer signals, extending \method to multi-modal and tool-augmented settings, and developing theoretical frameworks to understand when weak-to-strong succeeds. These directions highlight \method as a scalable and practical approach for eliciting strong reasoning capabilities in large language models.

\section{More Related Works}
\label{app:related}
\subsection{Chain-of-Thought Distillation}
Chain-of-thought (CoT) distillation has emerged as an effective technique for transferring reasoning abilities from large language models to smaller ones. Early work, such as Symbolic CoT Distillation (SCoTD)\cite{li2023symbolic}, showed that small models can benefit from CoT supervision generated by significantly larger teachers. Subsequent studies introduced methods like Keypoint-based Progressive CoT Distillation (KPOD)\cite{feng2024keypoint}, which incorporates token-level weighting and progressive learning to improve distillation. Other approaches, including CODI~\cite{shen2025codi} and DLCoT~\cite{luo2025deconstructing}, aim to compress CoT into continuous representations or deconstruct long reasoning trajectories for more efficient learning.

These methods typically rely on high-quality CoT traces from strong teacher models. In contrast, our work investigates the potential of leveraging structurally coherent but imperfect CoT traces from significantly weaker teachers to elicit strong reasoning capabilities in student models.
\subsection{Imperfect or Noisy Supervision}
Learning from imperfect or noisy supervision is a longstanding challenge in machine learning. In the context of LLMs, recent studies have examined the impact of noisy rationales on reasoning performance. For instance, Zhou et al.~\cite{zhou2024robust} proposed contrastive denoising methods to enhance robustness against noisy CoT prompts. Guo et al.~\cite{guo2024reliability} introduced reliability-aware alignment techniques to improve weak-to-strong generalization by assessing the trustworthiness of weak supervision signals.

Our work complements these efforts by demonstrating that even structurally flawed CoT traces from weaker models can effectively supervise stronger student models, highlighting the underexplored potential of leveraging imperfect yet structurally informative reasoning traces.

\section{Prompt Details}
\label{app:prompt}

Following~\cite{zeng2025simplerl}, for models with weaker instruction-following capabilities, such as Qwen-2.5-0.5B/1.5B, we adopt simpler prompts in previous work~\cite{abel}, which require only step-by-step reasoning. 
In contrast, for models with stronger instruction-following abilities, we employ more complex prompts as proposed by~\cite{qwen2.5}, where final answers must be explicitly placed within boxes.

\begin{table*}[h!]
    \small
    {\ttfamily
    \begin{tabularx}{\linewidth}{X}
    \toprule
    Question:\\
    \{input\} \\
    Answer: \\
    Let's think step by step. \\
    \bottomrule
    \end{tabularx}
    }
    \caption{Simple prompts template.}
    
    {\ttfamily
    \begin{tabularx}{\linewidth}{X}
    \toprule
    <|im\_start|>system \\
    You are a helpful assistant.<|im\_end|> \\
    <|im\_start|>user \\
    \{input\} \\
    Please reason step by step, and put your final answer within \textbackslash\textbackslash{}boxed\{\}.<|im\_end|> \\
    <|im\_start|>assistant \\
    \{output\} \\
    \bottomrule
    \end{tabularx}
    }
    \caption{Complex prompts template.}
    \label{tab:prompt}
\end{table*}

\section{Case Study}
\label{app:case}

As shown in~\cref{tab:case-1}, the weak teacher produces a structurally valid but numerically incorrect solution. The \method student retains this structured reasoning while correcting the computational errors, ultimately arriving at the correct answer. Its reasoning depth increases significantly, from 307 tokens (base student) to 984 tokens, approaching the length and quality of the RL-trained student (1038 tokens). These findings suggest that \method effectively leverages flawed yet structured supervision to elicit strong reasoning capabilities.

\begin{table*}[!h]
\caption{Case comparison of cognitive behaviors: weak teacher vs. base, RL, and \method students.}
    \centering
    \ttfamily
    \small
        \begin{tabular}{lp{11.5cm}}
        \toprule[1pt]
        \bf{Question} & 
Let $K$ be the number of sequences $A_1$, $A_2$, $\dots$, $A_n$ such that $n$ is a positive integer less than or equal to $10$, each $A_i$ is a subset of \{1, 2, 3, $\dots$, 10\}, and $A_{i-1}$ is a subset of $A_i$ for each $i$ between $2$ and $n$, inclusive. For example, \{\}, \{5, 7\}, \{2, 5, 7\}, \{2, 5, 7\}, \{2, 5, 6, 7, 9\} is one such sequence, with n = 5.What is the remainder when K is divided by 10?\\
        \midrule
        \bf{Weak Teacher} & 

To determine the number of sequences ( $A_1$, $A_2$, $\dots$, $A_n$ ) where n is a positive integer less than or equal to 10, each $A_i$ is a subset of \{1, 2, 3, dots, 10\}, and $A_{i-1}$ is a subset of $A_i$ for each i between 2 and n, inclusive, we need to consider the properties of such sequences.

Each element in the set \{1, 2, 3, dots, 10\} can either be in $A_i$  or not, and the choice for each element is independent of the others. Moreover, the condition $A_{i-1} \subseteq A_i$ means that if an element is in $A_i$, it must also be in $A_{i-1}$ for all i.

This problem can be translated into a combinatorial one. For each of the 10 elements, we need to decide in how many ways we can include or exclude it in the subsets $A_1$, $A_2$, $\dots$, $A_n$ such that the subset relationship is maintained. This is equivalent to choosing a non-decreasing sequence of subsets, which can be thought of as choosing a subset of \{1, 2, $\dots$, 10\} and then deciding how many times each element appears in the sequence.

For each element, there are n+1 choices (it can appear 0 times, 1 time, 2 times, ..., up to n times). Since there are 10 elements and each element's inclusion can be independently chosen, the total number of such sequences is given by:

$K = (n+1)^{10}$

We need to find the value of K for n ranging from 1 to 10 and then find the remainder when the sum of these values is divided by 10. However, since n can be any integer from 1 to 10, we need to sum $n+1)^{10}$ for n from 1 to 10:

$K = \sum_{n=1}^{10} (n+1)^{10}$

This simplifies to:

$K = 2^{10} + 3^{10} + 4^{10} + \cdots + 11^{10}$

We need to find the remainder of this sum when divided by 10. We can do this by finding the remainder of each term when divided by 10 and then summing these remainders.

$2^{10} \equiv 24 \equiv 4 \pmod{10}$

$3^{10} \equiv 59049 \equiv 9 \pmod{10}$

$4^{10} \equiv (4^2)^5 \equiv 16^5 \equiv 6^5 \equiv 7776 \equiv 6 \pmod{10}$

$5^{10} \equiv 0 \pmod{10}$

$6^{10} \equiv 6 \pmod{10}$

$7^{10} \equiv (7^2)^5 \equiv 49^5 \equiv 9^5 \equiv 59049 \equiv 9 \pmod{10}$

$8^{10} \equiv (8^2)^5 \equiv 64^5 \equiv 4^5 \equiv 1024 \equiv 4 \pmod{10}$

$9^{10} \equiv (9^2)^5 \equiv 81^5 \equiv 1^5 \equiv 1 \pmod{10}$

$10^{10} \equiv 0 \pmod{10}$

$11^{10} \equiv 1^{10} \equiv 1 \pmod{10}$

Summing these remainders:

4 + 9 + 6 + 0 + 6 + 9 + 4 + 1 + 0 + 1 = 40

The remainder when 40 is divided by 10 is:

$40 \equiv 0 \pmod{10}$

Thus, the remainder when K is divided by 10 is: $\boxed{0}$\\
\bottomrule[1pt]
    \end{tabular}
    \label{tab:case-1}
\end{table*}

\begin{table*}[!h]
    \centering
    \ttfamily
    \small
        \begin{tabular}{lp{11.5cm}}
        \toprule[1pt]
\bf{Base Student} & 
        To solve this problem, we need to count the number of sequences of subsets that satisfy the given conditions. Each subset $A_i$ must be a subset of $A_{i+1}$, and each subset must be a subset of \{1, 2, 3, dots, 10\}. This means that we can think of each element in \{1, 2, 3, dots, 10\} as having three possible states: it can be in none of the subsets, it can be in some of the subsets but not in all of them, or it can be in all of the subsets. For each element, there are 3 choices. Since there are 10 elements, there are $3^{10}$ possible sequences of subsets. The remainder when K is divided by 10 is the same as the remainder when $3^{10}$ is divided by 10. We can calculate $3^{10}$ using Python and then find the remainder when it is divided by 10. Let's do this calculation.
        
```python

\# Calculate $3^10$

K = 3**10

\# Find the remainder when K is divided by 10

remainder = K \% 10

print(remainder)

```

```output

9

```
The remainder when K is divided by 10 is $\boxed{9}$.\\
        \midrule
        \bf{RL Student} & 
To find the number of sequences  $A_1, A_2, \dots, A_n$ such that n  is a positive integer less than or equal to 10, each $A_i$  is a subset of \{1, 2, 3, $\dots$, 10\}, and $A_{i-1}$ is a subset of $A_i$  for each i between 2 and n , inclusive, we need to count the number of ways to choose such sequences for each possible value of n.

For a fixed  n , we need to count the number of ways to choose a chain of subsets $A_1 \subseteq A_2 \subseteq cdots \subseteq A_n \subseteq$ \{1, 2, $\dots$, 10\}. Each element in \{1, 2, $dots$, 10\} can independently choose to be in none of the sets, in exactly one of the sets, in exactly two of the sets, and so on, up to being in all  n  sets. Therefore, each element has  n+1 choices (it can be in 0, 1, 2, ..., or n  of the sets).

Since there are 10 elements, and each element has n+1 choices, the total number of such sequences for a fixed n is $(n+1)^{10}$.

We need to sum this over all possible values of n from 1 to 10:

$K = \sum_{n=1}^{10} (n+1)^{10}$

We are asked to find the remainder when  K  is divided by 10. Therefore, we need to compute:

$K \mod 10 = \sum_{n=1}^{10} (n+1)^{10} \mod 10$

We will compute each term $(n+1)^{10} \mod 10$ for n from 1 to 10:

$2^{10} = 1024 \equiv 4 \mod 10, $

$3^{10} = (3^2)^5 = 9^5 \equiv (-1)^5 = -1 \equiv 9 \mod 10, $

$4^{10} = (4^2)^5 = 16^5 \equiv 6^5 \equiv 6^2 \cdot 6^2 \cdot 6 \equiv 6 \cdot 6 \cdot 6 = 36 \cdot 6 \equiv 6 \cdot 6 = 36 \equiv 6 \mod 10, $

$5^{10} = (5^2)^5 = 25^5 \equiv 5^5 \equiv 5 \mod 10, $

$6^{10} = (6^2)^5 = 36^5 \equiv 6^5 \equiv 6 \mod 10, $

$7^{10} = (7^2)^5 = 49^5 \equiv (-1)^5 = -1 \equiv 9 \mod 10, $

$8^{10} = (8^2)^5 = 64^5 \equiv 4^5 \equiv 4^2 \cdot 4^2 \cdot 4 \equiv 6 \cdot 6 \cdot 4 = 36 \cdot 4 \equiv 6 \cdot 4 = 24 \equiv 4 \mod 10, $

$9^{10} = (9^2)^5 = 81^5 \equiv 1^5 = 1 \mod 10, $

$10^{10} = 10^5 cdot 10^5 \equiv 0^5 \cdot 0^5 = 0 \mod 10, $

$11^{10} = (11^2)^5 = 121^5 \equiv 1^5 = 1 \mod 10.$

Summing these results:

4 + 9 + 6 + 5 + 6 + 9 + 4 + 1 + 0 + 1 = 45

The remainder when 45 is divided by 10 is:

$45 \mod 10 = 5$

Therefore, the remainder when  K is divided by 10 is $\boxed{5}$.\\
        \bottomrule[1pt]
    \end{tabular}
\end{table*}

\begin{table*}[!h]
    \centering
    \ttfamily
    \small
        \begin{tabular}{lp{11.5cm}}
        \toprule[1pt]
        \bf{W2SR Student} & 

To solve for the number of sequences $A_1, A_2, dots, A_n$ where each $A_i$  is a subset of  \{1, 2, 3, dots, 10\}  and  $A_{i-1} \subseteq A_i$  for each  i  between 2 and  n, we need to consider the following:

1. **Understanding the Sequence**: Each sequence starts with the empty set $A_1$ = \{\} and ends with some subset $A_n$ of \{1, 2, 3, dots, 10\}. The key property is that each set in the sequence is a subset of the next set.

2. **Counting the Sequences**: For each element in \{1, 2, 3, dots, 10\}, we need to decide at which step it first appears in the sequence. If an element appears at step k, it must be in all subsequent sets $A_k, A_{k+1}, \dots, A_n$.

3. **Binary Decision for Each Element**: Each of the 10 elements can first appear at any step from 1 to n. Therefore, for each element, there are n choices (it can first appear at any of the n steps or not appear at all, which is the 0th step if we consider the empty set).

4. **Total Number of Sequences**: Since there are 10 elements and each element has n choices, the total number of sequences is $n^{10}$.

5. **Summing Over All Possible n**:  n can range from 1 to 10. Therefore, we need to sum $n^{10}$ for n from 1 to 10:

$K = sum_{n=1}^{10} n^{10}$
   
6. **Calculating the Sum**: We need to compute the sum of the 10th powers of the first 10 positive integers:

$K = 1^{10} + 2^{10} + 3^{10} + \cdots + 10^{10}$

7. **Finding the Remainder When K is Divided by 10**: We only need the last digit of  K. Therefore, we compute the last digit of each term $n^{10}$ for n from 1 to 10:

   $1^{10} = 1 \quad (\text{last digit is } 1)$
   
   $2^{10} = 1024 \quad (\text{last digit is } 4)$
   
   $3^{10} = 59049 \quad (\text{last digit is } 9)$
   
   $4^{10} = 1048576 \quad (\text{last digit is } 6)$ 
   
   $5^{10} = 9765625 \quad (\text{last digit is } 5)$
   
   $6^{10} = 60466176 \quad (\text{last digit is } 6)$ 
   
   $7^{10} = 282475249 \quad (\text{last digit is } 9)$
   
   $8^{10} = 1073741824 \quad (\text{last digit is } 4)$
   
   $9^{10} = 3486784401 \quad (\text{last digit is } 1)$
   
   $10^{10} = 10000000000 \quad (\text{last digit is } 0)$

   8. **Summing the Last Digits**:

   1 + 4 + 9 + 6 + 5 + 6 + 9 + 4 + 1 + 0 = 45

   The last digit of 45 is 5.
   
   Therefore, the remainder when K is divided by 10 is ($\boxed{5}$).\\
        \bottomrule[1pt]
    \end{tabular}
\end{table*}

\end{document}